\documentclass[runningheads]{llncs}

\usepackage{graphicx}
\usepackage{amsmath}
\usepackage{url}
\usepackage{siunitx}

\newcommand{\br}{\boldsymbol{r}}
\DeclareMathOperator*{\argmax}{arg\,max}
\newcommand{\netname}{PV-SynthSeg} 

\begin{document}


\title{Partial Volume Segmentation of Brain \\ MRI Scans of any Resolution and Contrast}
\titlerunning{Partial Volume Segmentation of MRI Scans of any Resolution and Contrast}

\author{Benjamin Billot \inst{1} \and
Eleanor Robinson \inst{1} \and
Adrian V. Dalca\inst{2,3}\and \\
Juan Eugenio Iglesias\inst{1,2,3}}

\authorrunning{Billot, Robinson, Dalca and Iglesias}

\institute{Centre for Medical Image Computing, University College London, United Kingdom
\and Martinos Center for Biomedical Imaging, Massachusetts General Hospital and Harvard Medical School, USA
\and Computer Science and Artificial Intelligence Laboratory, Massachusetts Institute of Technology, USA}

\maketitle 

\vspace{-0.3cm}
\begin{abstract}

Partial voluming (PV) is arguably the last crucial unsolved problem in Bayesian segmentation of brain MRI with probabilistic atlases. PV occurs when voxels contain multiple tissue classes, giving rise to image intensities that may not be representative of any one of the underlying classes. PV is particularly problematic for segmentation when there is a large resolution gap between the atlas and the test scan, e.g., when segmenting clinical scans with thick slices, or when using a high-resolution atlas. Forward models of PV are realistic and simple, as they amount to blurring and subsampling a high resolution (HR) volume into a lower resolution (LR) scan. Unfortunately, segmentation as Bayesian inference quickly becomes intractable when ``inverting'' this forward PV model, as it requires marginalizing over all possible anatomical configurations of the HR volume. In this work, we present \netname{}, a convolutional neural network (CNN) that tackles this problem by directly learning a mapping between (possibly multi-modal) LR scans and underlying HR segmentations. \netname{} simulates LR images from HR label maps with a generative model of PV, and can be trained to segment scans of any desired target contrast and resolution, even for previously unseen modalities where neither images nor segmentations are available at training. \netname{} does not require any preprocessing, and runs in seconds. We demonstrate the accuracy and flexibility of our method with extensive experiments on three datasets and 2,680 scans. The code is available at  \url{https://github.com/BBillot/SynthSeg}.

\keywords{Partial volume \and image segmentation \and brain MRI}

\vspace{0.3cm}
Published at the International Conference on Medical Image Computing and Computer-Assisted Intervention (MICCAI) 2020

\end{abstract}


\section{Introduction}
 
Segmentation of brain MRI scans is a key step in neuroimaging studies, as it is a prerequisite for an array of subsequent analyses, e.g., volumetry or connectivity studies. Although manual segmentation remains the gold standard, this expensive procedure can be replaced by automated tools, which enable reproducible segmentation of large datasets. However, a well-known problem of automated segmentation is the partial volume (PV) effect~\cite{choi_partial_1991,niessen_multiscale_1999}. PV arises when different tissues are mixed within the same voxel during acquisition, resulting in averaged intensities that may not be representative of any of the underlying tissues. For instance, in a T1 scan, the edge between white matter and cerebrospinal fluid (CSF) will often appear the same color as gray matter, even though no gray matter is present. This problem particularly affects scans with low resolution in any orientation (e.g., clinical quality images with thick slices), and fine-detailed brain regions like the hippocampus in research quality scans.
 
Modern supervised segmentation approaches based on convolutional neural networks (CNN)~\cite{kamnitsas_efficient_2017,milletari_v-net_2016,ronneberger_u-net_2015} can learn to segment with PV, given appropriate training data. However, they do not generalize well to test scans with significantly different resolution or intensity distribution~\cite{akkus_deep_2017,jog_psacnn_2019,karani_lifelong_2018}, despite recent advances in transfer learning and data augmentation~\cite{chaitanya_semi-supervised_2019,eaton-rosen_improving_2018,jog_psacnn_2019,long2017deep,shin2016deep,zhao_data_2019}. In contrast, Bayesian segmentation methods stand out for their generalization ability, which is why they are used by all major neuroimaging packages (e.g., FreeSurfer~\cite{fischl_freesurfer_2012}, SPM~\cite{ashburner_unified_2005}, and FSL~\cite{patenaude_bayesian_2011}). Bayesian segmentation with probabilistic atlases builds on generative models that combine a prior describing neuroanatomy (an atlas) and a likelihood distribution that models the image formation process (often a Gaussian mixture model, or GMM, combined with a model of bias field). Bayesian inference is used to ``invert'' this generative model and compute the most likely segmentation given the observed intensities and the atlas. Unfortunately, these models can  be greatly affected by PV.

A popular class of Bayesian methods uses an unsupervised likelihood term and estimates the GMM parameters from the test scan, which makes them adaptive to MRI contrast~\cite{ashburner_unified_2005,puonti_fast_2016,van_leemput_automated_1999,zhang_segmentation_2001}. This is a highly desirable feature in neuroimaging, since differences in hardware and pulse sequences can have a large impact on the accuracy of supervised approaches, which are not robust to such variability. Unsupervised likelihood models also enable the segmentation of \emph{in vivo} MRI with high resolution atlases built with \emph{ex vivo} modalities (e.g., histology~\cite{iglesias_probabilistic_2018}).

PV can easily be incorporated into the generative model of Bayesian segmentation by considering a high resolution (HR) image generated with the conventional non-PV model, and by appending smoothing and subsampling operations to yield the observed low resolution (LR) image. However, inferring the most likely HR segmentation from the LR voxels quickly becomes intractable, as estimating the model parameters requires to marginalize over the HR label configurations. Early methods attempted to circumvent this limitation by approximating the posterior of the HR label~\cite{laidlaw_partial-volume_1998,nocera_robust_1997}, or by explicitly modeling the most common PV classes (e.g., white matter with CSF) with dedicated Gaussian intensity distributions~\cite{noe_partial_2001,shattuck_magnetic_2001}. Van Leemput et al.~\cite{van_leemput_unifying_2003} formalized the problem and proposed a principled statistical framework for PV segmentation. They were able to simplify the marginalization and solve it for simple cases, given specific assumptions on the number of mixing classes and blurring kernel. Even with these simplifications, their method remains impractical for most real world scans, particularly when multiple MRI contrasts with different resolutions are involved.

In this paper, we present \netname{}, a novel and fast method for PV-aware segmentation of (possibly multi-modal) brain MRI scans. Specifically, we propose to synthesize training scans based on the forward  model of Bayesian segmentation, with a focus on PV effects. We train a CNN with these scans, which are generated on the fly with random model parameters~\cite{billot_learning_2020}. The CNN can be  trained to segment scans of any desired target resolution and contrast by adjusting the probability distribution of these parameters. As with classical Bayesian segmentation, the method only needs segmentations (no images) as training data. \netname{} leverages machine learning to achieve, for the first time, PV segmentation of MRI scans of unseen, arbitrary resolution and contrast without any limiting simplifying assumptions. \netname{} is very flexible and can readily segment multi-modal and clinical images, which would be unfeasible with exact Bayesian inference.


\section{Methods}

\subsection{Generative model of MRI scans with PV: intractable inference}
\label{bayesian}

Let $A$ be a probabilistic atlas that provides, at each spatial location, a vector with the occurrence probabilities for $K$ neuroanatomical classes. The atlas is spatially warped by a deformation field $\phi$ parametrized by $\theta_\phi$, which follows a distribution $p(\theta_\phi)$. Further, let $L=\lbrace L_{j}\rbrace _{1\leq j \leq J}$  be a 3D label map (segmentation) of $J$ voxels defined on a HR grid, where  $L_{j} \in \lbrace 1,...,K\rbrace$. We assume that each $L_j$ is independently drawn from the categorical distribution given by the deformed atlas at each location:
$
p(L , \theta_\phi | A) = p(\theta_\phi ) p(L | \theta_\phi , A) = p(\theta_\phi ) \prod_{j=1}^J p(L_j | \theta_\phi , A). 
$

Given a segmentation $L$, image intensities $I=\lbrace I_{j}\rbrace _{1\leq j \leq J}$ at HR are assumed to be independent samples of a (possibly multivariate) GMM conditioned on the anatomical labels: 
$
p(I,\theta_G, \theta_B | L) = p(\theta_G) p(\theta_B) \prod_{j=1}^J \mathcal{N} \left( I_j - B_j(\theta_B) ; \mu_{L_j}, \Sigma_{L_j} \right),
$
where $\theta_G$ is a vector grouping the means and covariances associated with each of the $K$ classes, and $B_j(\theta_B)$ is the bias field at voxel $j$ in logarithmic domain, parameterized by $\theta_B$. Both $\theta_G$ and $\theta_B$ have associated prior distributions $p(\theta_G)$ and $p(\theta_B)$, which complete the classical non-PV model. 

We model PV by assuming that, instead of the HR image $I$, we observe $\mathcal{D}(I) = \lbrace \mathcal{D}(I)_{j'}\rbrace _{1\leq j' \leq J'}$, defined over a coarser LR grid with $J'<J$ voxels, where $\mathcal{D}$ is a blurring and subsampling operator. If the blurring is linear, the likelihood $p(\mathcal{D}(I) | L, \theta_B,\theta_G)$ is still Gaussian (since every LR voxel is a linear combination of Gaussian HR voxels) but, in general, does not factorize over $j'$.

Bayesian segmentation often uses point estimates for the model parameters to avoid intractable integrals. This requires finding the most likely model parameters given the atlas and observed image, by maximizing $p(\theta_\phi,\theta_B,\theta_G | \mathcal{D}(I),A)$. Applying Bayes' rule and marginalizing over the unknown segmentation, the optimization problem is:
$$
\argmax_{\theta_\phi,\theta_B,\theta_G} p(\theta_\phi) p(\theta_B) p(\theta_G)  \sum_L p( \mathcal{D}(I) | L, \theta_B,\theta_G) p(L | \theta_\phi, A).
$$

\noindent Without PV (i.e., $\mathcal{D}(I)=I$), the sum over segmentations $L$ is tractable because both the prior $p(L | \theta_\phi, A)$ and the likelihood $p( I | L, \theta_B,\theta_G)$ factorize over voxels. However, in the PV case, blurring introduces dependencies between the underlying HR voxels, and the sum is intractable, as it requires evaluating $K^J$ terms.
Even with simplifying assumptions, such as limiting the maximum number of classes mixing in a LR voxel to two, using a rectangular blurring kernel, and exploiting redundancy in likelihood computations~\cite{van_leemput_unifying_2003}, computing the sum is prohibitively expensive: it requires $K (K-1) 2^{(M-1)}$ evaluations of the prior and $K (K-1) (1+M) /2 $ evaluations of the likelihood (where $M$ is the voxel size ratio between LR and HR), and  only remains tractable for very low values of $M$.

\subsection{PV-aware segmentation with synthesis and supervised CNNs}

\begin{figure}[t!]
\begin{center}
\includegraphics[width=0.88\textwidth]{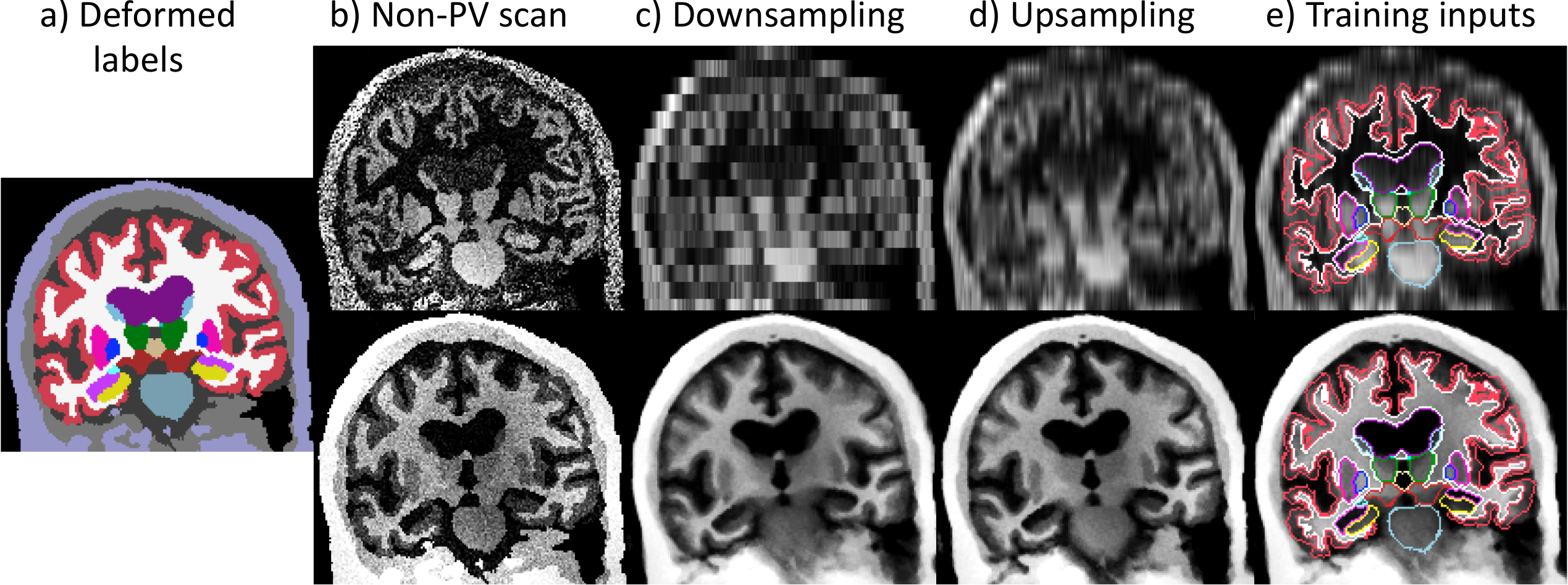}
\caption{Generation of a synthetic multi-modal MRI scan (1$\times$9$\times$\SI{1}{\milli\meter} axial FLAIR and a 1$\times$1$\times$\SI{9}{\milli\meter} coronal T1). We sample a HR image~(b) from a deformed label map~(a). We then simulate PV scans at LR with blurring and subsampling steps~(c). The LR scans are upscaled to the isotropic HR voxel grid~(d) to generate training pairs~(e).}
\label{steps}
\end{center}
\end{figure}

Rather than explicitly inverting the PV model of Bayesian segmentation, we employ a CNN that directly learns the mapping between LR intensities $\mathcal{D}(I)$ and HR labels $L$. We train this network with synthetic images sampled from the generative model (see example in Fig.~\ref{steps}). Specifically, every minibatch consists of a synthetic MRI scan and a corresponding segmentation, generated as follows. 

\textit{(a)}
Starting from a training dataset $\{S_t\}$ with $T$ segmentations, we first use a public GPU implementation~\cite{billot_learning_2020} to sample the non-PV joint distribution:
\begin{equation}
p(I,L,\theta_\phi,\theta_G,\theta_B | \{S_t\}) = p(I | L,\theta_G,\theta_B) p(L | \theta_\phi , \{S_t\}) p(\theta_\phi) p(\theta_G) p(\theta_B),
\label{eq:nonPVgenerativeModel}
\end{equation}
where the standard probabilistic atlas prior is replaced by a model where a label map is randomly drawn from $\{S_t\}$ and deformed with $\phi$, i.e., $ p(L | \theta_\phi , \{S_t\}) = (1/T) \sum_t \delta[L = (S_t \circ \phi)]$, where $\delta$ is the Kronecker delta. This model yields label maps that are more spatially regular than atlas samples (Fig.~\ref{steps}.a). The deformation field $\phi$ is obtained by sampling a stationary velocity field as a 10$\times$10$\times$10$\times$3  zero-mean Gaussian field with diagonal covariance, integrating it with a scaling-and-squaring approach~\cite{arsigny_log-euclidean_2006} to obtain a diffeomorphic field, and composing it with a random linear transform, with translation, rotation, scaling and shearing parameters sampled from uniform distributions. The intensity parameters $p(\theta_G)$ are sampled independently for each MRI contrast, using Gaussian distributions for the means and the logarithm of the variances. The bias field is obtained by sampling a 4$\times$4$\times$4 zero-mean Gaussian field with diagonal covariance, upscaling it to the input volume size, and taking the element-wise exponential. This process yields a multi-modal HR image~$I$ from a  HR label map~$L$ (Fig.~\ref{steps}.b).

\textit{(b)}
We simulate voxel thickness independently for each channel of $I$, by blurring them with anisotropic Gaussian kernels to simulate the target resolution of the LR images. Specifically, we design the standard deviation of the kernel such that the power of the HR signal is divided by 10 at the cut-off frequency. As the standard deviations in the spatial and discrete frequency domain are related by $\sigma_f \sigma_s = (2\pi)^{-1}$, the standard deviation of the blurring kernel is:
$$
\sigma_s  = 2  \log(10) / (2\pi)   \br_n / r_a \approx (3/4) \br_n / r_a,
$$
where $\br_n$ is the (possibly anisotropic) voxel size of the test scan in channel $n$, and $r_a$ is the isotropic voxel size of the atlas. We further multiply $\sigma_s$ by a factor $\alpha$ ($\sigma_s  = 0.75 \alpha  \br_n / r_a$), sampled from a uniform distribution of predefined range, to introduce small resolution variations and increase robustness in the method.

\textit{(c)}
Because in real data slice thickness and spacing are not necessarily equal, we simulate slice spacing by subsampling the blurred version of $I$ (still defined in the HR grid) to obtain $\mathcal{D}(I)$, defined on the LR grid (Fig.~\ref{steps}.c). 

\textit{(d)}
Finally, we upsample $\mathcal{D}(I)$ back into the original HR space with linear interpolation (Fig.~\ref{steps}.d). This step mimics the processing at test time, when we upscale the input to the target isotropic HR, so that the CNN can obtain a label map on the HR grid that represents anatomy within the LR voxels.

\subsection{Learning}

We train a 3D U-net~\cite{ronneberger_u-net_2015} with synthetic pairs generated on the fly with the PV model. The U-net has 5 levels with 2 layers each (3$\times$3$\times$3 kernel size and ELU activation~\cite{clevert_fast_2016}). The first layer has 24 kernels, this number being doubled after each max-pooling, and halved after each upsampling. The last layer uses a softmax activation. The optimization loss is defined as the average soft Dice coefficient over all predicted labels~\cite{milletari_v-net_2016}. Our method (generative model and CNN) is entirely implemented on the GPU, using Keras~\cite{chollet_keras_nodate} with a Tensorflow backend~\cite{abadi_tensorflow_2016}.

The hyperparameters governing the distributions of $\theta_\phi$ and $\theta_B$ are drawn from uniform distributions with relatively wide ranges (Table~S1 in the supplementary material), which increases the robustness of the CNN~\cite{billot_learning_2020}. The hyperparameters of $\theta_G$ are modality specific. In practice, we estimate them from unlabeled scans as follows. First, we run a publicly available Bayesian segmentation method (SAMSEG~\cite{puonti_fast_2016}). Second, we compute estimates of the means and variances of each class using robust statistics (median and median absolute deviation). Importantly, the estimated variances are multiplied by the ratio of the voxel size volumes at HR and LR, such that the blurring decreases the variances to the expected levels at LR. And third, we fit a Gaussian distribution to these parameters. Finally, we artificially increase the estimated standard deviations by a factor of 5, with two purposes: making the CNN resilient to changes in acquisition parameters, and mitigating segmentation errors made by SAMSEG.


\section{Experiments and results}

\subsection{Datasets}

\noindent\textbf{T1-39:} 39  1 mm isotropic T1 brain scans with segmentation for 39 regions of interest (ROIs)~\cite{fischl_freesurfer_2012}: 36 cerebral (manual) and 3 extra-cerebral (semi-automated). 

\smallskip
\noindent\textbf{FLAIR}: 2413 T2-FLAIR scans from ADNI~\cite{alzheimers} at 1$\times$1$\times$\SI{5}{\milli\meter} resolution (axial).

\smallskip
\noindent\textbf{CobraLab:} 5 multimodal (T1/T2) .6 mm isotropic scans~\cite{winterburn_novel_2013} with manual labels for 5 hippocampal subregions (CA1, CA23, CA4, subiculum, molecular layer). We segmented the rest of brain ROIs with FreeSurfer to obtain dense label maps.

\smallskip
\noindent\textbf{ADNI-HP:} 134 Alzheimer's disease (AD) cases and 134 controls from ADNI~\cite{alzheimers}, with T1 (1 mm) and T2 (.4$\times$.4$\times$2 mm coronal, covering only the hippocampus).

\subsection{Experimental setup}
\label{exp set up}

We evaluate \netname{} with three sets of experiments:

\smallskip
\noindent\textbf{T1-spacing:} We assess performance at different PV levels with the T1-39 dataset. We simulate sparse clinical scans in coronal, sagittal and axial orientation, at 3, 6 and \SI{9}{\milli\meter} slice spacing, with 3 mm slice thickness. We use our method to train a network to provide segmentations on the 1 mm isotropic grid. We use segmentations from 20 cases for training, and the rest of the subjects for testing.

\smallskip
\noindent\textbf{FLAIR:} To evaluate our method on scans representative of clinical quality data, with real thick-slice images and a contrast other than T1, we use the same 20 label maps from the T1-39 dataset to train our method to segment the FLAIR scans, on a 1 mm isotropic grid. The Gaussian hyperparameters are estimated from a subset of 20 FLAIR scans, and the remaining 2393 are used for testing. We use FreeSurfer~\cite{fischl_freesurfer_2012} segmentations of corresponding  T1 ADNI scans as ground truth. We emphasize that such T1 scans are often not available in clinical protocols, but here we can use these for evaluation purposes only.

\smallskip
\noindent\textbf{Hippocampus:} We also evaluate our method on a multi-modal MRI dataset with different resolutions for each channel, in the context of a neuroimaging group study. We use the segmentations from the CobraLab dataset to train our model to segment the hippocampal subregions on the ADNI-HP dataset, on the \SI{0.6}{\milli\meter} isotropic grid. Since no ground truth is available for the target dataset, we use the ability to separate groups and detect known atrophy patterns in AD~\cite{fox_presymptomatic_1996,jack_prediction_1999,yushkevich2010nearly,mueller2018systematic} as a proxy for segmentation accuracy.  

\smallskip
We compare the proposed approach with two other competing methods. First, Bayesian segmentation without PV; this is a natural alternative to our approach, as it only requires label maps for supervision, and adapts to MRI contrast (including multi-modal). In the first two experiments, we use SAMSEG~\cite{puonti_fast_2016} (trained on the same 20 scans from T1-39) to segment the upsampled HR inputs (we also tried segmenting the LR scans directly, with inferior results). In the third experiment, we use a  publicly available hippocampal segmentation algorithm~\cite{iglesias2015computational}, with a probabilistic atlas created from the CobraLab data. 

The second competing approach is a supervised CNN trained on LR images from the target modality, which requires paired imaging and segmentation data. We test this approach on the first and third experiments, which represent the settings in which manual labels may be available. Specifically, we train the same 3D U-net architecture with real scans blurred to the target resolution, and using the same augmentation strategy as for our method. We emphasize that such methods are only applicable in more rare supervised settings, but the performance of these networks provides an informative upper bound for the accuracy of \netname{}. We evaluate all methods on both the HR (``dense") and the LR grid (``sparse"), obtained by downsampling the HR labels.

\subsection{Results}

\begin{figure}[t!]
\begin{center}
\includegraphics[width=\textwidth]{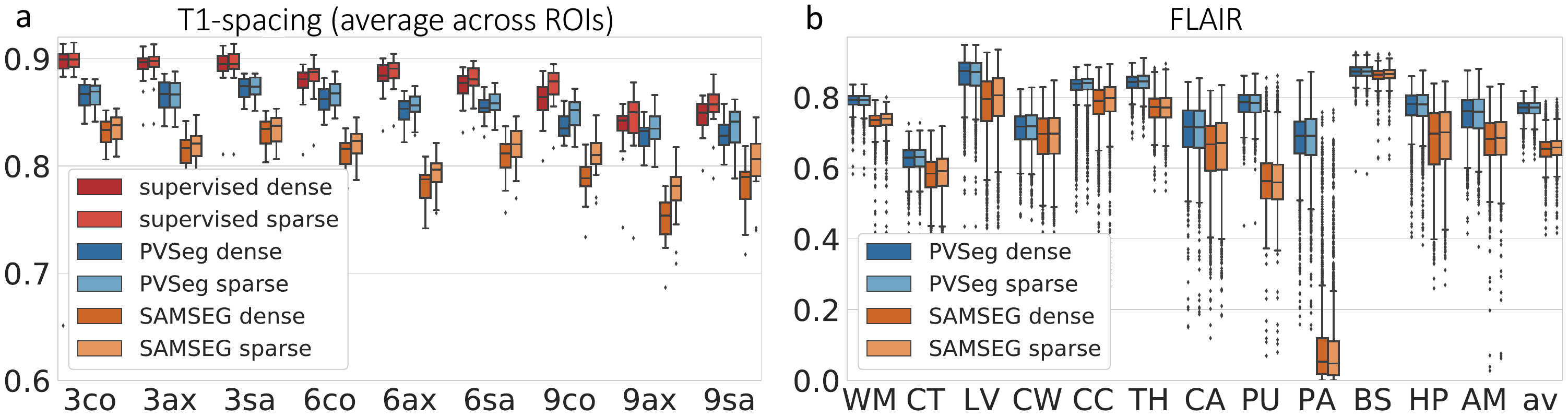}
\caption{(a) Box plot of Dice scores in T1-spacing experiment with  3, 6 and \SI{9}{\milli\meter}  spacing in coronal (co), axial (ax), and sagittal (sa) orientations, averaged over 12 representative ROIs: cerebral white matter (WM) and cortex (CT); lateral ventricle (LV); cerebellar white matter (CW) and cortex (CC); thalamus (TH); caudate (CA); putamen (PU); pallidum (PA); brainstem (BS); hippocampus (HP); and amygdala (AM). (b) Box plot of Dice scores  for the 12  ROIs in the FLAIR experiment and their average (av).}
\label{dice}
\end{center}
\end{figure}

\begin{figure}[t!]
\begin{center}
\includegraphics[width=\textwidth]{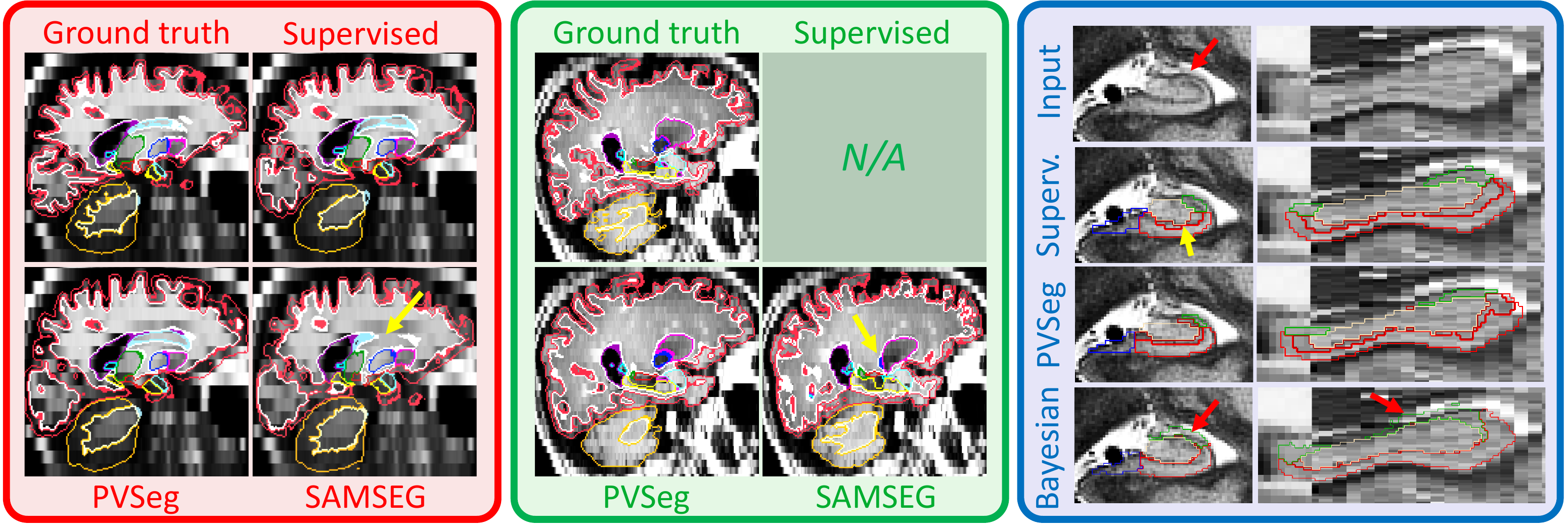}
\caption{Examples of dense segmentations. Red box:  1$\times$1$\times$\SI{9}{\milli\meter} volume from T1-spacing experiment. Green box: sample from FLAIR experiment. Blue box: Hippocampus  (T2 with partial coverage overlaid on T1). See main text for a description of the arrows. More sample segmentations are shown in Figures~S2-S4 in the supplement. }
\label{segmentations}
\end{center}
\end{figure}

Figure~\ref{dice}.a shows the mean Dice scores for the T1-spacing experiment. \netname{}  consistently outperforms SAMSEG by up to 6 Dice points, and is robust to large slice spacings: even at 9 mm, it yields competitive Dice scores (0.83 mean), both when evaluated densely and on the sparse slices. Comprehensive structure-wise results  are shown in Fig.~S1 in the supplement; they reveal that, with increasing slice spacing, accuracy decreases the most for the thin and convoluted cerebral cortex. This is also apparent from the example in Figure~\ref{segmentations} (red box,  1$\times$1$\times$9 mm resolution), where the cortex is inaccurate for all methods. Due to PV effects, SAMSEG almost completely fails to segment the caudate (yellow arrow), which our method successfully recovers. Having access to the exact intensity distributions, the supervised approach outperforms \netname{} (only marginally at higher slice spacing), but is only an option in the rare scenario where one has access to manually labeled HR scans of the target contrast.

\begin{table}[t]
\setlength\tabcolsep{4pt}
\caption{Effect sizes (Cohen's $d$) and $p$ values of non-parametric Wilcoxon tests, comparing the volumes of the hippocampal substructures in AD subjects vs. controls.}
\centering
\begin{tabular}{|c|c|c|c|c|c|c|}
\hline
Method & CA1 & CA23 & CA4 & Subiculum & Molec. Layer & Whole \\
\hline
Supervised: $d$ & 1.94 & 1.66  & 1.87 & 1.84 & 2.15 & 2.10 \\  
Supervised: $p$ & $< 10^{-29}$ & $< 10^{-23}$ & $< 10^{-28}$ & $< 10^{-27}$ & $< 10^{-34}$ & $< 10^{-33}$ \\ 
\hline
\netname{}: $d$ & 2.06 & 1.62 & 1.73 & 1.33 & 1.48 & 1.92 \\  
\netname{}: $p$ & $< 10^{-32}$ & $< 10^{-25}$ & $< 10^{-25}$ & $< 10^{-15}$ & $< 10^{-19}$ & $< 10^{-29}$ \\ 
\hline
Bayesian: $d$ & 1.93 & 1.42 & 1.73 & 1.96  & 0.48 & 1.79 \\  
Bayesian: $p$ & $< 10^{-29}$ & $< 10^{-18}$ & $< 10^{-24}$ & $< 10^{-29}$ & $< 10^{-4}$ & $< 10^{-26}$ \\ 
\hline
\end{tabular}
\label{tab:adni}
\end{table}

On FLAIR scans, \netname{} achieves a mean Dice score of 0.77 (Fig.~\ref{dice}.b). This is a remarkable result, considering the low contrast of these scans and their large slice thickness (5 mm). In contrast, SAMSEG only yields 0.65 Dice (12 points below), and consistently labels the pallidum as putamen. This is shown in Figure~\ref{segmentations} (green box), where the pallidum is pointed by the yellow arrow. 
Although \netname{} uses hyperparameters computed with SAMSEG, it successfully recovers the pallidum (Dice $\approx$ 0.75), which highlights its robustness against inaccurate hyperparameter estimation. \netname{} is also noticeably more accurate in other structures in this example, like the hippocampus (in yellow). 
As in
T1-spacing, 
neither method is accurate for the cortex at this resolution (Dice $\approx$ 0.60) -- again, partly due to the low gray-white matter contrast.

In the hippocampus experiment, PV effects in the T2 scan cause the interface between white matter and the lateral ventricle to appear as gray matter, leading to segmentation errors in the Bayesian algorithm (red arrows in blue box of Figure~\ref{segmentations}). Despite having been trained on only five cases, the supervised method does not have this problem, but follows the internal structure of the hippocampus (revealed by the molecular layer: the dark band pointed by the yellow arrow) much less accurately than \netname{}. While all three methods detect large effect sizes in the AD experiment (Table~\ref{tab:adni}), \netname{} replicates well-known differential atrophy patterns (derived from manual~\cite{mueller2018systematic} and semi-automated segmentations~\cite{yushkevich2010nearly}) much better than the other two approaches, with CA1 showing stronger atrophy than CA4, and the subiculum  remaining relatively spared.


\section{Conclusion}

We have presented \netname{}, a novel learning-based segmentation method for brain MRI scans with PV effects. \netname{} can accurately segment most brain ROIs in scans with very large slice thickness, regardless of their contrast (even when previously unseen), and replicates differential atrophy patterns in the hippocampus in an AD study. A general limitation of PV segmentation is the low accuracy for the cortex at larger spacing, which precludes application to cortical thickness and parcellation analyses. We will tackle this problem by combining our approach with image imputation. \netname{} enables morphology studies of large clinical datasets of any modality, which has enormous potential in the discovery of imaging biomarkers in a wide array of neurodegenerative disorders.


\section*{Acknowledgement}

Work supported by the ERC (Starting Grant 677697), EPSRC (UCL CDT in Medical Imaging, EP/L016478/1), Alzheimer’s Research UK (Interdisciplinary Grant ARUK-IRG2019A-003), NIH (1R01AG064027-01A1, 5R01NS105820-02), the Department of Health's NIHR-funded Biomedical Research Centre at UCLH.


\bibliographystyle{splncs04}


\newpage
\renewcommand{\thetable}{S\arabic{table}}
\makeatother
\makeatletter

\title{Supplementary Materials: Billot et al. \\ Partial Volume Segmentation of MRI Scans}
\titlerunning{Partial Volume Segmentation of MRI Scans of any Resolution and Contrast}
\author{}
\authorrunning{Billot, Robinson, Dalca and Iglesias}
\maketitle 
\vspace{-10ex}

\begin{center}
\includegraphics[width=0.96\textwidth]{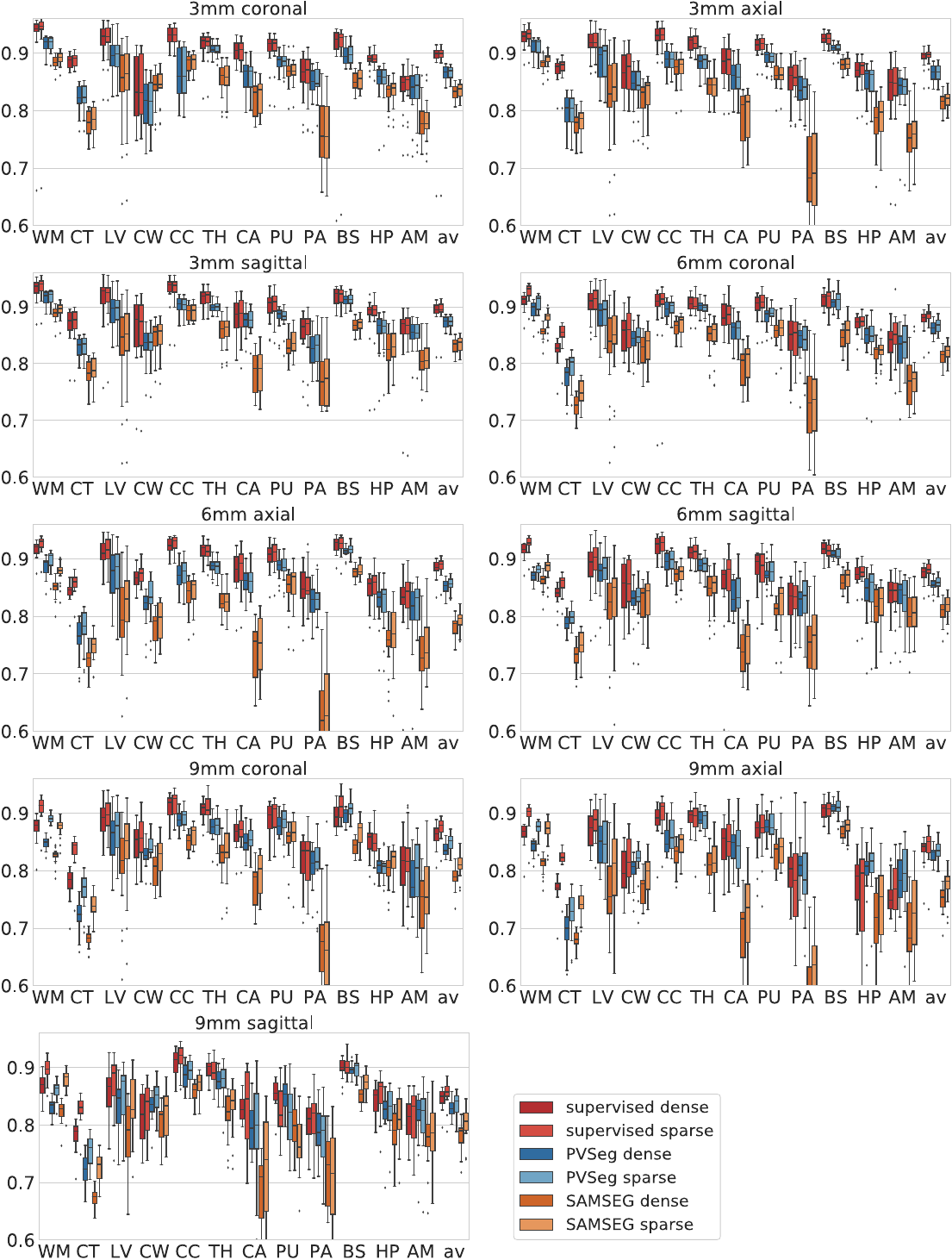}
\end{center}
\textbf{Figure S1.} Structure-wise box plots of Dice scores in T1-spacing experiment for each spacing and orientation, including 12  ROIs and their average (av): cerebral white matter (WM) and cortex (CT); lateral ventricle (LV); cerebellar white matter (CW) and cortex (CC); thalamus (TH); caudate (CA); putamen (PU); pallidum (PA); brainstem (BS); hippocampus (HP); and amygdala (AM).

\begin{center}
\includegraphics[width=\textwidth]{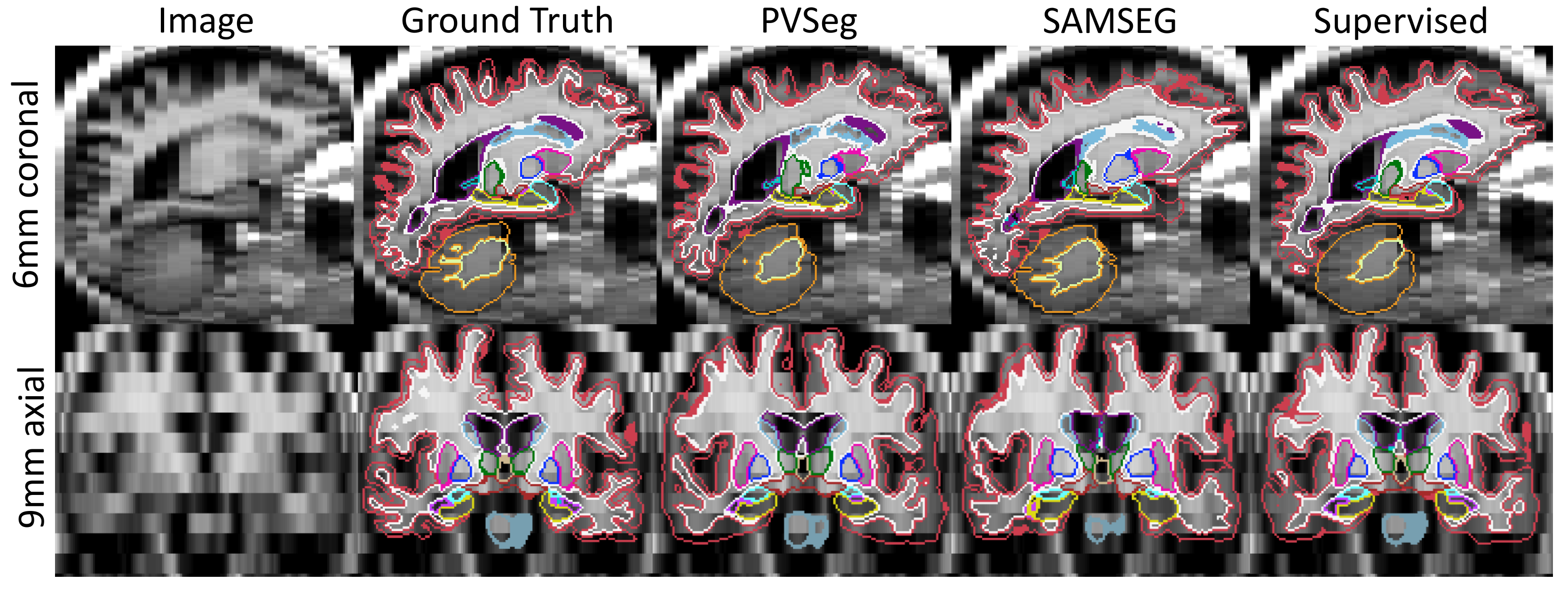} \\
\textbf{Figure S2.} Two more examples from T1-spacing experiment.
\end{center}

\begin{center}
\includegraphics[width=0.80\textwidth]{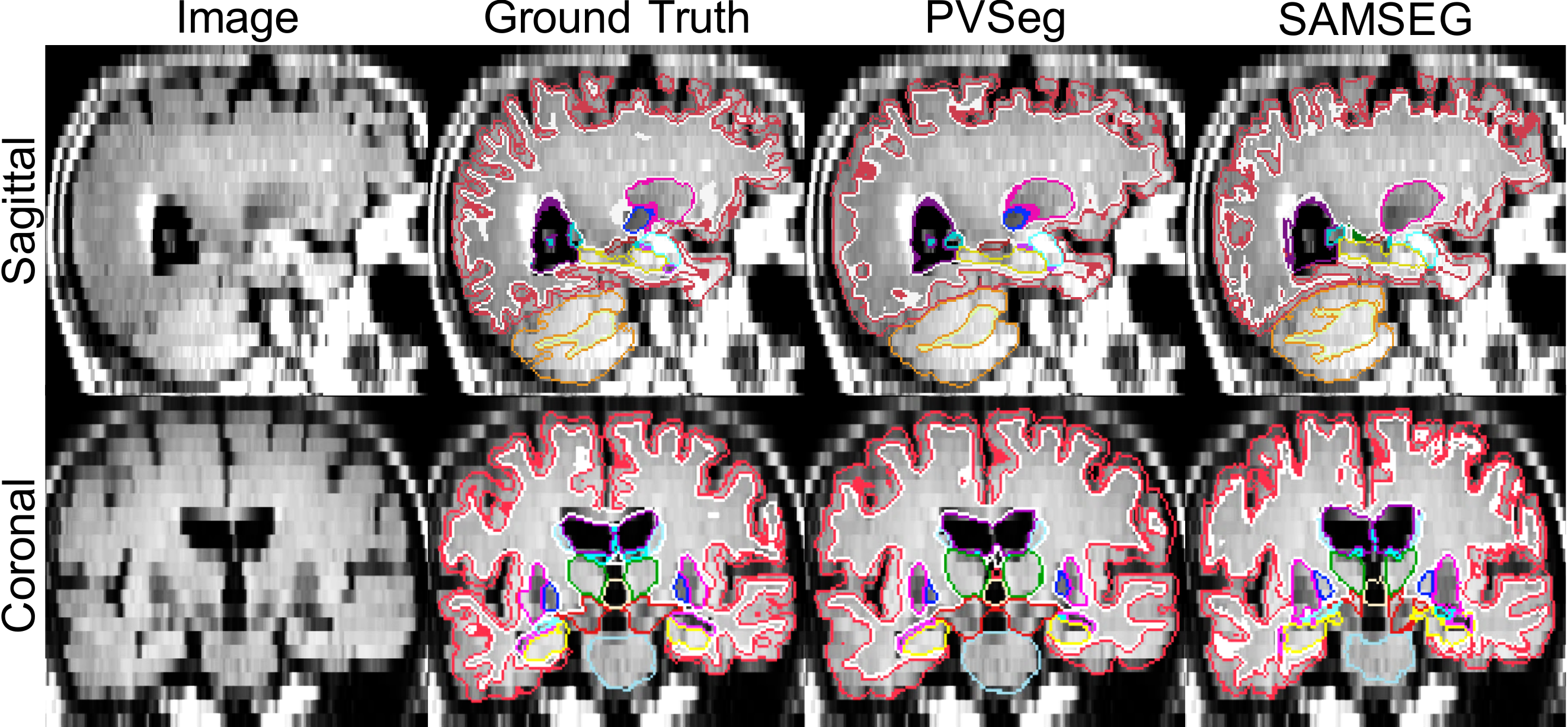} \\
\textbf{Figure S3.} Two more examples of FLAIR segmentations.
\end{center}

\begin{center}
\includegraphics[width=0.925\textwidth]{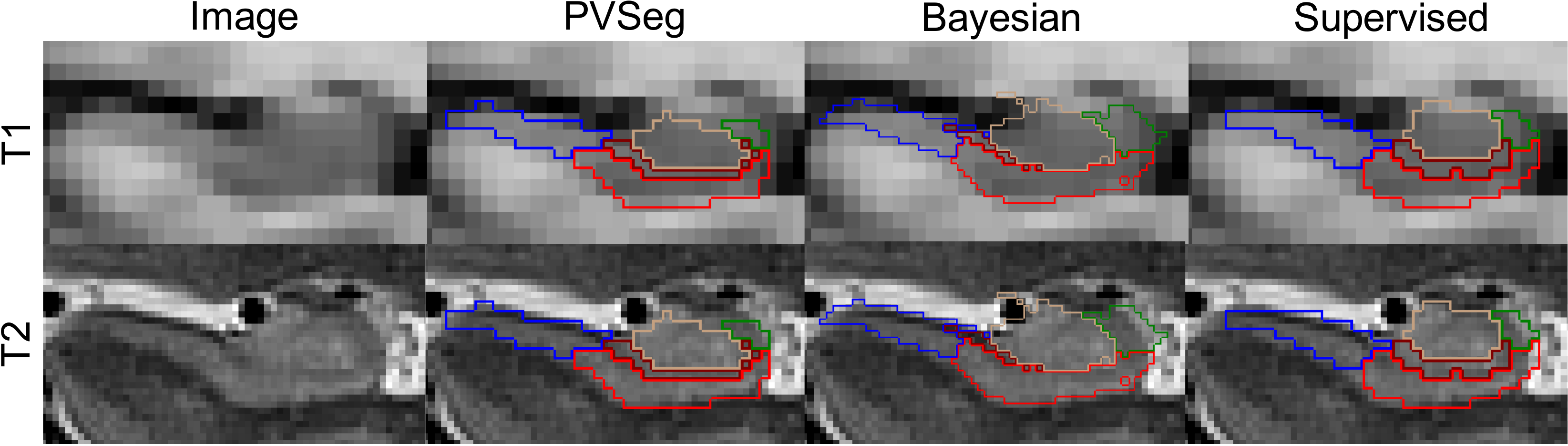} \\
\textbf{Figure S4.} Close-up of coronal view of co-registered T1 (1 mm isotropic) and T2 scan (0.4$\times$0.4$\times$2.0 mm) of a sample case from the hippocampus dataset.
\end{center}

\begin{table}[b]
\setlength\tabcolsep{7.5pt}
\caption{Ranges of the uniform distributions for the parameters of the generative model: rotation ($\theta_{rot}$); scaling ($\theta_{sc}$); shearing ($\theta_{sh}$); translation ($\theta_{tr}$); standard deviation for generation of the stationary vector field ($\sigma_v$) and bias field ($\sigma_b$); and factor for the blurring kernel that simulates voxel thickness ($\alpha$).}
\centering
\begin{tabular}{|c|c|c|c|c|c|c|c|}
\hline
$\theta_{rot} \ (^\circ)$ & $\theta_{sc}$ & $\theta_{sh}$ & $\theta_{tr}$ & $\sigma_v$ & $\sigma_b$ & $\alpha$ \\
\hline
[-15, 15] & [0.8, 1.2] & [-0.01, 0.01] & [-20, 20] & [0, 4] & [0, 0.5] & [0.75, 1.25] \\
\hline
\end{tabular}
\end{table}

\end{document}